# Advancing the Scientific Method with Large Language Models: From Hypothesis to Discovery

Yanbo Zhang[1,*+], Sumeer A. Khan[2,6,7], Adnan Mahmud[3,4], Huck Yang[5], Alexander Lavin[8], Michael Levin[1,9], Jeremy Frey[10], Jared Dunnmon[11], James Evans[12,13], Alan Bundy[14], Saso Dzeroski[15], Jesper Tegner[2,16,17,18,19,*+], Hector Zenil[20,21,22,23,24*+]

1. Allen Discovery Center at Tufts University, Medford, MA, USA
2. Living Systems Lab, KAUST, Thuwal, Kingdom of Saudi Arabia
3. Intelligent Infrastructure Team, Network Rail, UK
4. Department for AI in Society, Science, and Technology, Zuse Institute Berlin, Germany
5. NVIDIA Research
6. Biological and Environmental Science and Engineering Division, King Abdullah University of Science and Technology, Thuwal, Saudi Arabia.
7. SDAIA-KAUST Center of Excellence in Data Science and Artificial Intelligence, Thuwal 23952, Saudi Arabia
8. Pasteur Labs, Brooklyn, NY, USA
9. Wyss Institute for Biologically Inspired Engineering, Harvard University, Boston, MA, USA
10. Department of Chemistry, University of Southampton, University Road, Southampton, Hampshire, UK
11. Department of Biomedical Data Science, Stanford University, Stanford, CA, USA
12. Knowledge Lab, Department of Sociology, University of Chicago, IL, USA
13. Santa Fe Institute, NM, USA
14. School of Informatics, The University of Edinburgh, UK
15. Department of Knowledge Technologies, Jožef Stefan Institute, Ljubljana, Slovenia
16. Biological and Environmental Science and Engineering Division, King Abdullah University of Science and Technology (KAUST), Thuwal 23955-6900, Saudi Arabia
17. Unit of Computational Medicine, Department of Medicine, Center for Molecular Medicine, Karolinska Institutet, Karolinska University Hospital, L8:05, SE-171 76, Stockholm, Sweden
18. Computer, Electrical and Mathematical Sciences and Engineering Division, King Abdullah University of Science and Technology (KAUST), Thuwal 23955-6900, Saudi Arabia.
19. Science for Life Laboratory, Tomtebodavagen 23A, SE-17165, Solna, Sweden
20. Algorithmic Dynamics Lab, Research Departments of Biomedical Computing and Digital Twins, School of Biomedical Engineering and Imaging Sciences, King's College London, UK
21. King's Institute for Artificial Intelligence, London, UK
22. The Alan Turing Institute, London, UK
23. Oxford Immune Algorithmics, Oxford University Innovation and London Institute for Healthcare Engineering, London, UK
24. Cancer Interest Group, Francis Crick Institute, London, UK

* Corresponding author: hector.zenil@kcl.ac.uk

## Abstract

With recent Nobel Prizes recognising AI contributions to science, Large Language Models (LLMs) are transforming scientific research by enhancing productivity and reshaping the scientific method. LLMs are now involved in experimental design, data analysis, and workflows, particularly in chemistry and biology. However, challenges such as hallucinations and reliability persist.

In this contribution, we review how Large Language Models (LLMs) are redefining the scientific method and explore their potential applications across different stages of the scientific cycle, from hypothesis testing to discovery. We conclude that, for LLMs to serve as relevant and effective creative engines and productivity enhancers, their deep integration into all steps of the scientific process should be pursued in collaboration and alignment with human scientific goals, with clear evaluation metrics.

The transition to AI-driven science raises ethical questions about creativity, oversight, and responsibility. With careful guidance, LLMs could evolve into creative engines, driving transformative breakthroughs across scientific disciplines responsibly and effectively. However, the scientific community must also decide how much it leaves to LLMs to drive science, even when associations with 'reasoning', mostly currently undeserved, are made in exchange for the potential to explore hypothesis and solution regions that might otherwise remain unexplored by human exploration alone.

# Introduction

Recent advances in artificial intelligence (AI) have transformed multiple areas of society, the world economy, and academic and scientific practice. Generative AI and Large Language Models (LLMs) present unprecedented opportunities to transform scientific practice, advance Science, and accelerate technological innovation. Nobel Prizes in Physics and Chemistry were awarded to several AI leaders for their contributions to AI and frontier models, such as Large Language Models (LLMs). This promises to transform or contribute to scientific research by enhancing productivity and supporting various stages of the scientific method. The use of AI in science is booming across numerous scientific areas and is impacting different parts of the scientific method.

Despite the potential of LLMs for hypothesis generation and data synthesis, AI and LLMs face challenges in fundamental science and scientific discovery. Hence, our premise in our perspective is that AI, in general, has so far been limited in its impact on fundamental science, which is defined here as the discovery of new principles or new scientific laws. Here, we review how LLMs are currently used – as a technological tool – to augment the scientific process in practice and how they may be used in the future as they become more powerful tools and develop into powerful scientific assistants. Combining data-driven techniques with symbolic systems, such a system could fuse into hybrid engines that may lead to novel research directions. We aim to describe the gap between LLMs as technical tools and "creative engines" that could enable new high-quality scientific discoveries and pose novel questions and hypotheses to human scientists. We first review the current use of LLMs in Science, aiming to identify limitations that need to be addressed when moving toward creative engines.

There is solid recognition and excitement for the transformative potential of AI in Science. For example, leading machine learning conferences (NeurIPS, ICML) have recently (2021-2023) arranged targeted workshops on AI4Science. Some recent reviews and papers include[1–38]. This demonstrates the energy and potential of using automated (i.e., AI tools) for Science. This "dream" can be traced back to the times of Turing and the emergence of Artificial Intelligence in the 1950s[39]. With recent advancements in computational techniques, vastly increased production of scientific data, and the rapid evolution of machine learning, this long-held vision can be transformed into reality. Yet, most current reviews and original papers focus on specifically designed machine learning architectures targeting particular application domains or problems.

For example, recent reviews have explored how to use variants of Deep Learning, Geometric Deep Learning, or Generative AI in its generality (including different architectures such as CNNs, GNNs, GANs, diffusion models, VAEs, and Transformers) as a tool for assisting Science [3,11,13,15,19,22]. For example, Wang et al.[1], reviews breakthroughs in how specific techniques such as geometric deep learning, self-supervised learning, neural operators, and language modelling have augmented Science in protein folding, nuclear fusion, and drug discovery. An essential thread in their review is the vital notion of representation, pointing out that different AI architectures can support valuable representations of scientific data and thereby augment Science. Recent papers demonstrate the appeal and the potential of using AI-driven and augmented tools for automating science[1,4,13,40]. Traditional scientific advancements have been primarily driven by hypothesis-led experimentation and theoretical development, often limited by human cognitive capacities and manual data processing. For example, the formulation of Newtonian mechanics required meticulous observation and mathematical formalization over extended periods. Here, the rise of AI4Science represents a paradigmatic revolution that could reach beyond human cognitive limitations. AI-driven advancements promise to enable rapid processing and analysis of massive data sets, revealing complex patterns that surpass human analytical capabilities. For example, DeepMind's AlphaFold dramatically transformed protein structure prediction, a longstanding scientific challenge, using deep learning to predict protein folding accurately. Furthermore, AI4Science could reverse the slowdown in scientific productivity in recent years, where literature search and peer-review evaluation[41–43] are bottlenecks.

In contrast to previous reviews, here we first address the use of LLMs, regardless of the specific underlying architecture, and their use as a tool for the scientific process. We assess how different areas of science use LLMs in their respective scientific process. This analysis sets the stage for asking how LLMs can synthesize information, generate new ideas and hypotheses, guide the scientific method, and augment fundamental scientific discoveries. Here, we ask to what extent AI can be described as a "general method of invention," which could open up new paradigms and directions of scientific investigations.

Hence, complementary to a purely representational and architectural viewpoint of AI4Science, we find it constructive to ask and assess to what extent the nature of the scientific process, both its inductive and deductive components, can and should be transformed by AI techniques.

## Current use of LLMs – From Specialised Scientific Copilots to LLM-assisted Scientific Discoveries

The ability of Large Language Models (LLMs) to process and generate human-like text, handle vast amounts of data, and analyse complex patterns with potentially some reasoning capabilities has increasingly set the stage for them to be used in scientific research across various disciplines. Their applications range from simple tasks, such as acting as copilots to assist scientists, to complex tasks, such as autonomously performing experiments and proposing novel hypotheses. We will first introduce the fundamental concepts of LLMs and then review their various applications in scientific discovery.

### Prompting LLMs: From Chatbot to Prompt Engineering

Current mainstream LLMs are primarily conditional generative models, where the input, such as the beginning of a sentence or instructions, serves as a condition, and the output is the generated text, such as a reply. This text is typically sampled auto-regressively: the next token (considered the building block of words) is sampled from a predicted distribution. See Figure 1A.

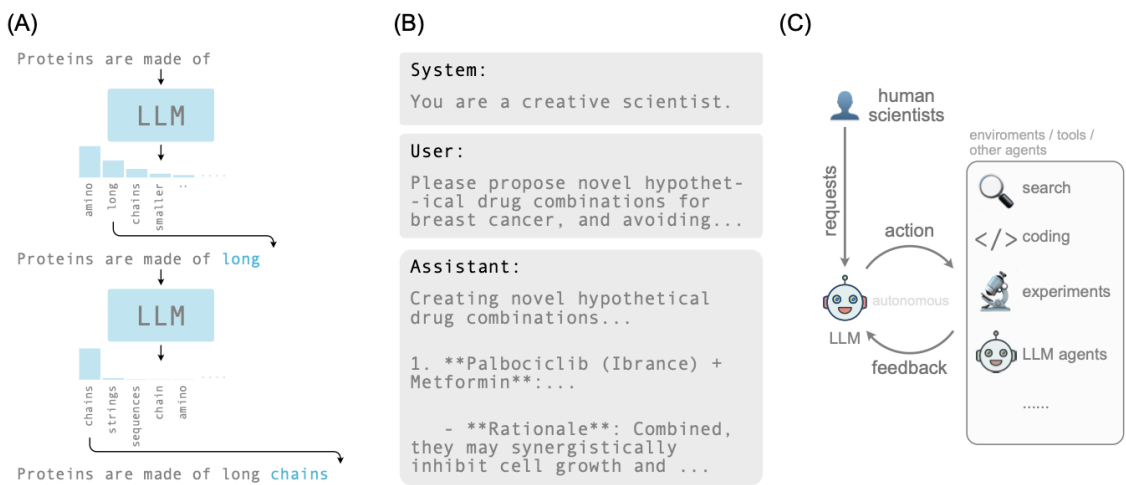

Figure 1: (A) LLMs generate sentences in an auto-regressive manner, sampling tokens from a predicted distribution at each step. (B) A typical prompt for LLMs consists of a system prompt and a user prompt. The LLM will then respond as an assistant. A multi-round dialogue will repeat the user and assistant contents. (C) LLM agents are systems that use a large language model as its core reasoning and decision-making engine, enabling it to interpret instructions, plan actions, and autonomously interact with external tools, environments, or other LLM agents to fulfil a given goal.

Given LLMs' capabilities in computation and emerging potential for reasoning, which we define as the ability to solve tasks that require reasoning, they can be considered programming languages that use human language as the code that instructs them to perform desired tasks. This code takes the form of "prompts." For instruct-tuned LLMs, the prompt often consists of three parts: the system prompt and the user prompt, with an LLM's reply considered the assistant prompt. Hence, a chat is frequently composed of **<system><user><assistant><user><assistant>**, see Figure 1B. The system prompt typically includes general instructions for the LLMs, such as behaviour, meta-information, format, etc. The user prompt usually contains detailed instructions and questions. Using these prompts, the LLMs generate replies under the role of "assistant.

Since LLMs do not have background knowledge about the user, and prompts are their major input, designing a good prompt is often critical to achieving the desired output and superior performance. Researchers have shown that specific prompts, including accuracy, creativity, and reasoning, can significantly improve output performance.

Specifically, the chain-of-thought (CoT) method[44] can instruct LLMs to think step-by-step, leading to better results. Beyond these, the Retrieval-augmented Generation (RAG) method[45] can incorporate a large amount of context by indexing the contents and retrieving relevant materials, then combining the retrieved information with prompts to generate the output. Due to the importance of prompts and LLM agents, designing prompts is now often called "prompt engineering," and many techniques and tricks have been developed in this area[46,47], such as asking JSON format outputs, formulating clear instructions, setting temperatures, etc[47,48].

While carefully designed prompts can accomplish many tasks, they are not robust and reliable enough for complex tasks requiring multiple steps or non-language computations, nor can they explore autonomously. LLM agents are developed for these requirements, especially for complex tasks. LLM agents are autonomous systems powered by LLMs, which can actively seek to observe environments, make decisions, and perform actions using external tools[49]. In many cases, we need to ensure reliability, achieve high-performance levels, enable automation, or process large amounts of context. These tasks cannot be accomplished solely with LLMs and require integrating LLMs into agent systems. Early examples include AutoGPT[50] and BabyAGI[51], where LLMs are treated as essential tools within the agent system (Figure 1C). In scientific discovery, LLM agents become even more critical due to the complexity of science and its high-performance requirements. Many tools have also been developed to provide easy access to these prompting and agent methods, such as LangChain[52] and LlamaIndex[53]. Automated prompt design methods, such as DSPy[54] and TextGrad[55], are also being developed to design prompts and LLM agents in a data-driven way.

**LLMs as Practical Scientific Copilots**

The ability of LLMs to work with a large body of text is being exploited in the practice of science. For example, LLMs assist in proposing novel ideas, writing scientific papers and generating computer code, thereby improving productivity; they also adapt texts for diverse audiences ranging from experts to broader audiences, thus supporting communication in science.

Furthermore, LLMs can sift through vast bodies of scientific literature to identify relevant papers, findings, and trends. Such reviewing of the relevant literature helps investigators quickly digest and identify gaps in enormous bodies of scientific knowledge. These capabilities can also mitigate discursive barriers across different scientific fields, supporting interdisciplinary scientific collaborations and knowledge sharing. Recently, chatbots have emerged in several disciplines as virtual assistants answering scientific queries posed by scientists. Such tools exploit the power of LLMs to extract and detect patterns, data, and knowledge. These techniques may also serve as important tools in science education and communication.

These examples demonstrate the rise of LLMs in extracting and sharing information and the exciting open research frontier of the potential of reasoning that they represent in different scientific domains[56–58]. For instance, Caufield et al. proposed the SPIRES method[59], which uses LLMs to extract structured data from the literature. Beyond data extraction, LLMs have also shown evidence of outperforming annotation tasks[60,61], enabling scientists to scale data annotation. Some domain-specific models also show superior performance in classification, annotation, and prediction tasks[62–64]. With the help of RAG methods[45], LLMs can directly apply their information extraction and distillation capabilities to large amounts of text data. With the combination of diverse capabilities of LLMs interconnected through LLM-agents, the recent "AI co-scientist[65]" demonstrates impressive ability in generating novel research ideas by leveraging existing literature, engaging in internal LLM-agent debates, and refining its outputs. This process leads to constructive progress when applied to real scientific tasks.

Moreover, LLMs are currently used to automate the experimental design and the execution of experiments. For example, Boiko, et al.[66] propose an autonomous LLM capable of performing chemical experiments. This work employs an LLM planner to manage the experimental process, such as drawing patterns on plates or conducting more complex chemical syntheses. Compared to hard-coded planners, the LLM-based planner is more flexible and can handle unexpected situations. Similar kinds of loop and tool usage are also shown in[67], which includes literature tools, consulting with humans, experimental tools, and safety tools.

In the biological domain, for instance, the CRISPR-GPT[68] represents a significant advancement in biological research. It utilizes LLMs to automate the design of gene-editing experiments, enhancing both the efficiency and precision of genetic modifications, which is pivotal in speeding up genomic research and applications. Another advance in the application of LLMs in the biological domain is BioDiscoveryAgent[69]. These tools augment scientists' capabilities and accelerate scientific discovery.

The capabilities described thus far capture the current use of LLMs as knowledge engines. Summarising, extracting, interfacing, and reasoning about (scientific) text, alongside automating experimental design and execution. While immensely useful, it remains an open frontier on how to do this safely and efficiently. It largely depends on how prompting is performed and how LLM agent systems are designed.

**Foundation Models for Science**

A key observation when using LLMs as clever text engines or exploiting the underlying machine learning (neural) architecture for solving specific scientific problems was the importance of scale. Larger models trained on larger amounts of data, or spending larger amounts of computation during inference time yielded an increase in performance[56,70,71]. The discovery of such scaling laws[72] demonstrated that LLMs' performance improves as the number of parameters increases. Thus, we can expect the above trends to grow in importance as these systems are trained on ever larger amounts of data. Emergent behaviours, such as reasoning were suggested when models increased in scale[73]. Concurrent with the appreciation of scaling laws came the realisation that instead of using LLMs for specialised problems or as text engines, one could potentially train them on large amounts of data, not necessarily text, but different modalities of scientific data. This is the idea of a foundation model. These are large-scale pre-trained models that, when trained with a sufficient amount of data of different types, such models "learn" or "encapsulate" knowledge of a large scientific domain, thus reaching beyond a specific scientific problem. When fine-tuned to particular tasks, such models can solve a wide range of downstream tasks. The notion of foundation models refers to their generality in that they can be adapted to many different applications, unlike task-specific engineered models solving a specialised task such as protein folding. Notably, the famous transformer architecture that fuels LLMs has become the architecture of choice when constructing the foundational models in different domains of science. These self-supervised models are usually pre-trained on extensive and diverse datasets. This enables them to learn from massive unlabelled data since masking parts of the data and then requiring the model to predict the occluded parts provides foundation models with their learning objective. This technique is used when training LLMs on large amounts of text. The idea is thus exploited in scientific domains where multi-modal data is used to train self-supervised foundation models. Once trained, the model can be fine-tuned for various downstream tasks without requiring additional training. Consequently, the same model can be applied to a wide range of downstream tasks. The foundation model encapsulates a large body of scientific "knowledge" inherent in the training data.

Leveraging these ideas, there has been a rise in the number of foundation models of science. For example, the Evo and Evo 2 models enable prediction and generation tasks from the molecular to the genome scale[74]. While Evo is trained on millions of prokaryotic and phage genomes, Evo 2[75] includes massive eukaryotic genomes, and both demonstrate zero-shot function prediction across DNA, RNA, and protein modalities. It excels at multimodal generation tasks, as shown by generating synthetic CRISPR-Cas molecular complexes and transposable systems. The functional activity of Evo-generated CRISPR-Cas molecular complexes and IS200 and IS605 transposable systems was experimentally validated, representing the first examples of protein-RNA and protein-DNA co-design using a language model. Similarly, scGPT is for learning single cell transcriptional data[76], ChemBERT encodes molecular structures as strings, which then can be used for different downstream tasks such as drug discovery and material science[77]. Similarly, OmniJet-α is the first cross-task foundation model in particle physics, enhancing performance with reduced training needs[78]. Additionally, multiple physics pretraining (MPP) introduces a task-agnostic approach to modelling multiple physical systems, improving predictions across various physics applications without extensive fine-tuning[79]. The LLM-SR[80] implements similar symbolic regression methods iteratively, generating and evaluating hypotheses, using the evaluation signal to refine and search for more hypotheses.

Incorporating diverse scientific data modalities, which represent different "languages" to interact with observations beyond natural language, is crucial. There are two major approaches emerging: 1) End-to-end training on domain-specific modalities: Models like ChemBERT[77] (using chemical SMILES strings) and scGPT[76] (using single-cell data), as mentioned above, are directly trained on these specialized data types. 2) Separate training with compositional capabilities: This involves training separate encoders for new modalities or enabling LLM agents to utilize tools that interact with these modalities. For instance, models like BiomedCLIP[81] connect biological images with natural language, while PaperCLIP[82] and AstroCLIP[83] link astronomical images and spectral data to textual descriptions. Furthermore, frameworks like ChemCrow[84] leverage the tool-using abilities of LLMs to connect with non-natural-language modalities, such as chemical analysis tools.

Yet, as with text-based LLMs, several challenges remain. These include potential biases in datasets, which can bias the performance and output of these models. Since science is mainly about understanding systems, the scale, and opaqueness of these models make interpretation a particularly challenging problem. Also, several observations, such as their capability for generalisation, multi-modality, and apparent emergent capabilities, have led to intense discussions at the research frontier on the extent to which these foundation models can reason within and beyond their training regimes. The text-based LLMs (or models incorporated with text modality) discussed above are constructed using these techniques. Examples include GPT-4 (OpenAI)[85], BERT (Bidirectional Encoder Representation from Transformers)[86], CLIP (Contrastive Language-Image Pre-training, OpenAI)[87], and DALL-E from OpenAI[88].

These foundation models have the potential to achieve professional human-level performance or even surpass human capabilities when trained using reinforcement learning, particularly with feedback from reliable formal systems. For example, AlphaProof[89] has become state-of-the-art in automated theorem-proving systems, achieving mathematical capabilities comparable to human competitors at IMO 2024. Approximately one million informal mathematical problems were translated into the formal language LEAN, a mathematical proof verification language, enabling the LLM to be trained through reinforcement learning. Solutions generated by the LLM in LEAN are either proved or disproved by the LEAN compiler, with the resulting correct or incorrect solutions serving as feedback to refine the LLM. While this approach has been explicitly applied within the mathematical domain, it demonstrates significant potential for training LLMs to surpass human performance in highly complex and deductive reasoning. Although developing formal systems for general tasks remains challenging, reinforcement learning methods are employed to build foundation models with enhanced deductive capabilities, leading to the rise of reasoning models such as OpenAI o1/o3[70], Deepseek R1[56], and others. In scientific domains such as physics, external and reliable feedback mechanisms are already used to improve answer quality[90], highlighting the potential for creating domain-specific foundation models.

In conclusion, the rise of foundation models will continue to affect and disrupt science due to their powerful nature, scaling properties, and ability to handle very different data modalities. However, for our purposes, the question remains of what extent foundation models could be a proper gateway to making fundamental scientific discoveries. To what extent can foundation models be creative and reason outside their training domains?

## Toward Large Language Models as Creative Engines for Fundamental Science

Here, we ask how AI can impact fundamental Science? That is, what is required for an AI to be able to discover new principles of scientific laws from observations, available conjectures, and data analysis? Broadly, can generative AI develop to become a "creative engine" that can make fundamental scientific discoveries and pose new questions and hypotheses? Einstein famously stated, "*If I had an hour to solve a problem, I'd spend 55 minutes thinking about the problem and 5 minutes thinking about solutions*". This underscores the importance of carefully considering the question or problem itself, as posing hypotheses effectively can be the most intellectually demanding part of Science. As a first approximation, the ability to pose novel hypotheses is – at least for us humans – what appears to be essential for making novel discoveries. Thus, what is required for an AI to advance beyond a valuable tool for text generation and engineered systems for solving a particular problem? Or could foundation models provide a possible path forward?

In our view, if LLMs are to contribute to fundamental Science, it is necessary to assess what putative roles LLMs can play in the core of the scientific process. To this end, we discuss below how LLMs can augment the scientific method. This includes how LLMs could support observations, automate experimentation, and generate novel hypotheses. We will also explore how human scientists can collaborate with LLMs.

**Augmenting the Scientific Method**

As a first approximation, scientific discovery can be described as a reward-searching process, where scientists propose hypothetical ideas and verify or falsify them through experiments[91]. Under this Popperian formulation, LLMs can assist scientific discovery in two ways (Figure 2): On the one hand, LLMs could assist in the hypothesis-proposing stage, helping scientists find novel, valuable, or potentially high-reward directions or even propose hypotheses that human scientists might have difficulty generating. On the other hand, LLMs have the potential to make experiments more efficient, accelerate the search process, and reduce experimental costs.

At the stage of proposing hypotheses, scientists choose unknown areas to explore, which requires a deep command of domain knowledge, incorporating observational data, and manipulating existing knowledge in novel ways[45,92]. Their expertise and creativity could carry the potential for proposing novel research hypotheses.

Then, at the verification stage, experiments are conducted to obtain relevant information and test hypotheses. This requires the ability to plan and design experiments effectively. Given LLMs' planning capabilities and potential understanding of causality[93–95], they can help scientists design experiments. By incorporating tool-using abilities[96], LLMs can directly implement experiments. LLM agents can perform complex workflows and undertake repetitive explorations that are time-consuming for human scientists. This allows us to search for novel results efficiently, which is key to scientific discovery[97,98].

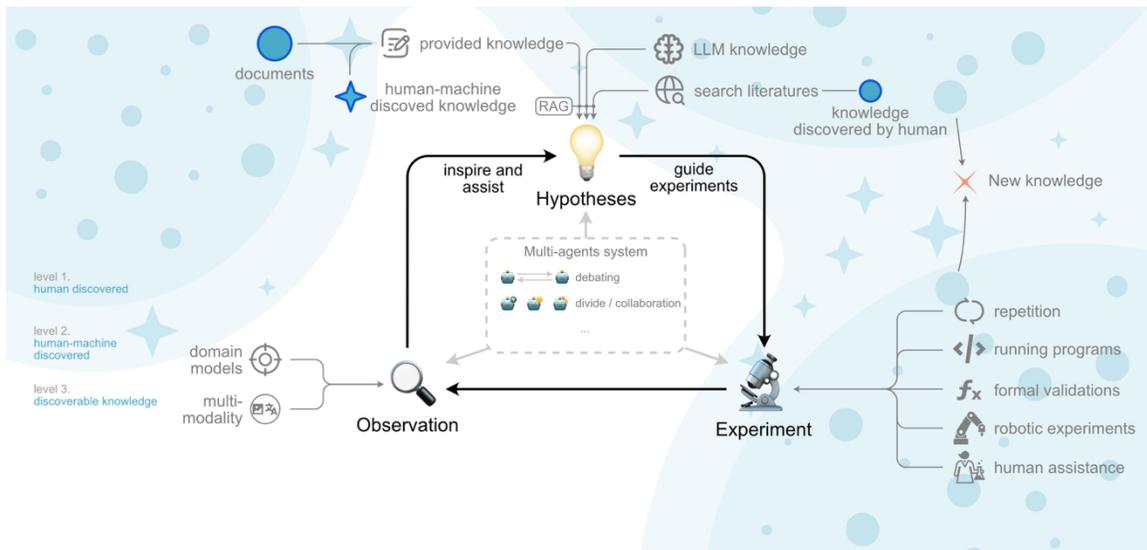

Figure 2: Illustration of the scientific discovery process: Scientific research can be formulated as a search for rewards in an abstract knowledge space. By synthesizing existing knowledge – represented by blue disks (human-discovered) and stars (human-machine discovered)) – in novel ways, new knowledge (indicated by red stars) can be explored. For specific research, scientists or LLMs need to traverse the hypotheses-experiment-observation loop, where hypotheses are proposed based on existing knowledge (including LLM knowledge, and additional literature provided via RAG methods), observation, and the creativity of LLMs. Then, with aid of external tools such as programming languages, formal validations, and other methodologies, experiments are conducted to test the hypotheses or gather data for further analysis. The experimental results can be observed and described through the observation process, facilitated by domain-specific models and the multi-modality capabilities of language models. All these parts–observation, proposing hypotheses, conducting experiments, and automation–can be assisted by LLMs and LLM-agents, considering the non-trivial implementations of scientific environments in silico.

This process often involves a trial-and-error loop for a research topic or question. Thus, scientific discovery requires the following steps: observation, hypothesis proposal, experimentation, and automating the loop.

**Expanding or Narrowing the Observation Process**

Scientists rely upon observational results for guidance in proposing hypotheses, designing and refining experiments, evaluating experimental results, and validating their hypotheses. In general, observations act as dimension reduction methods[99], which include annotating, classification, and information extraction.

General purpose LLMs, such as GPT-4, Llama, can be good observers for language and image data for general purposes. Their in-context and zero-shot learning capabilities can be used as universal classifiers to extract specific information from these data, such as annotation and evaluation. In domains like NLP and Social Science, annotating and evaluating language data at scale is a fundamental task for downstream experiments. Trained humans or crowd-workers have often done such jobs. However, LLMs, such as ChatGPT, can perform higher or comparable performance levels relative to crowd-workers on annotation tasks, especially on more challenging tasks[60,61].

Besides language processing, scientists must also describe complex behaviours at scale qualitatively. LLMs show potential in describing such complex black-box systems, where we observe only their inputs and outputs without knowledge of their underlying mechanisms. Although deciphering such systems can often become a stand-alone research question, having a qualitative description can still be helpful when faced with large-scale data. With LLMs, black-box systems, such as language input-output, mathematical input-output pairs, fMRI data[100], or observational data, can be described using natural language[100].

Beyond text and text-represented systems, different data modalities represent different "languages" to interact with observations, and domain-specific modalities are extremely important for scientific discovery. Scientific research often involves other data types, including image, video, audio, table[101,102], or even general files[103], as well as domain-specific modalities like genomic sequences, chemical graphs, or spectra[76,77,82,83]. Multi-modality LLMs can play the observer role vis-a-vis these data. However, most multi-modality LLMs are still struggling to handle some domain-specific data formats, such as genomic data or chemical compounds, which may require converting and where information may be lost during the conversion process.

For highly specialised domains, domain-specific LLMs trained on specialised data can achieve superior performance within their respective fields (representing the end-to-end approach discussed earlier). For example, with the same number of parameters, BioGPT[64] outperforms GPT-2 medium[104] when trained on domain-specific data. Even with fewer parameters, models like PubMedBERT[62] can perform at a level comparable to GPT-2 medium. In the chemical domain, LLMs have been pre-trained on chemical SMILES data[105], enabling them to infer molecular properties and biological activities[51]. LLM-inspired models are also useful for case-specific tasks. In[106], transformers are trained on cellular automata to study the relationship between complexity and intelligence. This highlights the importance of exploring domain-specific and case-specific LLM and the opportunities for further exploration in this area.

**Experimentation and Automation**

The experiment is a critical part of all research steps, including making observations and validating the hypothesis. Both humans and LLMs need external tools to implement experiments. Specifically, this involves calling external functions or directly generating and running code. LLMs that have been fine-tuned for tool usage[85,96] can generate structured entities (often in JSON) that contain the function name and inputs to be implemented by external functions. These functions are versatile and can include simple calculations, laboratory control functions, external memory, requests for assistance from human scientists, etc. LLMs can also direct programming by generating and running code for complex experiments requiring fine-grained control or enhancing the calculation abilities of LLMs[107,108]. Beyond this, generated programs can also call other functions or be saved into a function library, enabling the combinatory development of complex actions[109].

For complex experiments, planning becomes important, which involves setting up an objective and decomposing it into practical steps. This is critical to solving complex tasks while sustaining coherent behaviour. While the planning capabilities of LLMs are questioned in many studies, certain tools and methods still demonstrate valuable assistance. The chain-of-thought (CoT)[44] method significantly improves various tasks by decomposing a question into steps. In complex tasks with more steps, where LLMs seek long-term objectives and interact with environments, they can generate plans in natural language based on given objectives[110]. It is also important to adapt to observations and unexpected results. For this reason, methods like Reflexion[111], ReAct[112] combine the CoT and planning, dynamically update its plans, manage exceptions, and utilizes external information. And it also overcomes hallucination and error propagation in the chain-of-thought.

Automation is a significant aspect of LLM-assisted research, serving as a key contributor to accelerating scientific discovery. Automation involves repetition and feedback loops[113]. LLMs can be seen as a function -- prompt in, reply out – with human users as the driving force behind making LLMs produce output. To automate such a system, the key is to replace the human user. For instance, an LLM-powered chemical reaction system can perform Suzuki and Sonogashira reactions by incorporating an LLM-based planner, which replaces the human user. The planner reasons through the given task and determines the next steps, including searching the internet for information on both reactions, writing Python code to calculate experimental parameters, and finally calling external tools to conduct the experiments. At each step, the results, i.e., the search outcomes and calculation results, are fed back to the LLM-based planner to automate the system[66]. Another approach is to replace the human user with multiple LLMs and allow them to communicate with each other[114]. Since such automation is not fully hard-coded and the core of this automation is also an LLM, they can exhibit some emergent behaviour[66,114], adapting unexpected situations, which is vital for exploring new knowledge. Specifically, automated LLMs can help in three dimensions of scientific discovery: scaling, enhancing, and validation.

**Scaling:** Automated LLM agents can scale previously challenging experiments for large-scale studies. Examples include inferring underlying functions from input-output pairs[115]. The LLMs perform multiple rounds of trial and error to find the correct function. This approach can extend to neuron interpretation of GPT-2 using GPT-4, which has billions of parameters[116]. This method involves two layers of loops: the trial-and-error process and the application to all the billions of neurons[117,118]. Both layers are time-consuming for human scientists, and LLMs make such studies feasible. Another example is when LLMs are used to infer the functionality of human brain voxels from fMRI activation data, their proposed functions are first validated by calculating the probability of observing the activation data given a specific functional hypothesis. Subsequently, the hypotheses with the high probability are selected to aid in generating new hypotheses and improving overall performance[100]. Lab experiments can also be parallelized with the help of LLMs, which further accelerate the experiment speed and increase the potential for scaling the scope of experiments[110,119].

**Enhancing**: The aforementioned scientific methods, such as hypothesis generation, experiments, and observations, can all be enhanced by automation. One direct application is using LLMs as optimisers: by iteratively providing historical solutions and scores, LLMs can propose new solutions and ultimately achieve superior performance[120]. In both the hypothesis-validation loop and in experimental trials, failed cases constitute valuable feedback. When evaluators and reflection are incorporated into the workflow, LLMs can improve their decisions, showing significant performance improvements compared to simply using LLMs[111]. Iteration can also enhance the hypothesis generation stage. By comparing hypotheses with existing literature on related topics, LLMs can iteratively improve novelty by using this literature as a source of negative examples[121]. Another enhancement comes from accumulating knowledge, which is critical to research success. Many exploration tasks require accumulating knowledge and developing new strategies based on this knowledge[122]. For example, Voyager[109] uses GPT-4 to examine the space of the Minecraft game. This study consists of three main parts: an automatic curriculum to propose exploration objectives, an iterative prompting mechanism to write code to control the game, and a skill library to accumulate the knowledge and skills gained during the exploration, which is then reused in future explorations. Equipped with all these components, this LLM-assisted explorer can explore the game more efficiently. While game environments in silico are a non-trivial departure from real worlds in situ, they are not too dissimilar from the biochemical simulation engines[123] that scientists rely on today. However, the current "physics" engines in-game systems are still inconsistent with the physical sciences, and new simulation software technologies are needed to allow for any AI-based exploration of multi-physics environments[113]. From a macroscopic viewpoint, scientific discovery can also be considered a quality-diversity search process[124,125], and this Voyager study has shown how LLMs can assist diversity search in a new way by proposing objectives, iteratively solving problems, and contributing to and utilising literature (skill library).

**Validation**: Automated LLM agents are critical for validating hypothesis. Beyond scaling and enhancing performance, research often involves multiple rounds of the hypothesis experiment loop to meet scientific discovery's rigor and safety requirements. This loop is essential given the probabilistic nature of LLMs[126] and the hallucination problem of LLMs[127,128]. Experiments show that repeatedly verifying the results from LLMs' observations and proposed hypotheses increases the likelihood of obtaining reliable results[129,130]. A promising direction is leveraging formal systems to validate results and hypotheses by translating generated hypotheses and answers into formal languages, such as LEAN or Prover9[131,132]. For instance, in[131], LLMs first generate multiple answers. These answers are then translated into the LEAN language and verified using the LEAN compiler to choose the correct responses. With these filtered answers, LLMs can aggregate toward a final answer. Another example involves using Python code to aid validation. While general programming languages are often not considered formal systems, they can still disprove certain hypotheses. In[133], LLMs were prompted to solve the Abstraction and Reasoning Corpus (ARC) tasks[134], which involve identifying underlying laws and making predictions based on new initial states. LLMs initially propose hypotheses, which are then translated into Python code. This Python code is used to disprove incorrect hypotheses. Although these non-formal systems cannot fully validate hypotheses, they partially perform validation and improve predictive accuracy. While humans could also conduct such translation and validation processes, the high speed of hypothesis generation by LLMs makes automated approaches more suitable. A limitation, however, is the reliance on LLMs to translate hypotheses into formal languages, which may introduce errors in the process. This suggests the need for caution when interpreting results, even if they have been tested using formal systems.

**Expanding the Literature Review and the Hypothesis Horizon**

In brief, advancing beyond current knowledge includes using LLMs to explore unknown territories in knowledge space, encompassing human discoverable, human-machine discoverable, non-human-machine discoverable, and the entirety of the knowledge space, as illustrated in Figure 2. Namely, to perform hypothesis generation and develop predictive models of more complex systems. Hence, can LLMs do open-ended Exploration of the Hypothesis Space? Can LLMs also explore complex environments in an open-ended way? These are open-ended challenges addressing the (unknown) limits of the capabilities of LLMs and Generative AI.

Proposing hypotheses is a crucial step in scientific discovery, perhaps the most important since it often involves significant creativity and innovation. Scientists propose hypotheses to explore unknown topics or address research questions. This step often involves novel ideas, recombining existing literature, and key insights. Experiment design and subsequent verification are based on these hypotheses. Thus, hypothesis proposing is a central step that connects observation and experiments.

Evidence indicates that LLMs can propose novel ideas, such as drug combinations[135], with designed prompting, thus underscoring the importance of prompting, as discussed previously. An example is the use of LLMs for drug discovery: In[135] LLMs are prompted to propose novel combinations of drugs for treating MCF7 breast cancer cells while incorporating additional constraints such as avoiding harm to healthy cells and prioritizing FDA-approved and readily accessible drugs. The experiment results demonstrate that LLMs can effectively propose hypothetical drug combinations. More advanced techniques can also improve novelty, such as asking LLMs to role-play as scientists[136] or iteratively provide feedback on existing similar ideas[66]. This is further exemplified by the Virtual Lab project[137], where AI agents, powered by LLMs, were used to design novel nanobody binders against SARS-CoV-2 variants. LLMs effectively functioned as hypothesis generators, facilitating rapid and innovative scientific discovery that translates to validated experimental results in real-world applications. Although some human evaluations show that LLM-generated ideas have lower novelty[138], the fast speed at which LLMs propose ideas can still be valuable. With proper instruction and background knowledge, LLMs can act as zero-shot hypothesis generators[139]. LLMs can also generate hypotheses semantically or numerically based on observations about the underlying mechanisms for language processing and mathematical tasks[140,141]. With neuron activation heatmaps, GPT-4 can propose potential explanations for neuron behaviour[116].

Besides directly proposing hypotheses, a significant part of creativity combines existing knowledge, making literature research critical. With their vast stored compressed knowledge[142,143], LLMs can be viewed as databases queried using natural language[123]. This not only accelerates the search but also breaks down barriers of domain terminology, making it easier to access interdisciplinary knowledge. For accessing more up-to-date and domain-specific information, LLMs can help scientists by using the RAG method and accessing internet information, see Figure 2. Generally, text embedding is used for semantically searching vector databases[45,92]. For example, STORM[144] proposes an LLM-powered literature review agent that, for a given topic, actively searches literature on the internet from different perspectives and automatically generates follow-up questions to improve depth and thoroughness. Another important example is Deep Research[145,146], which integrates internet browsing and reasoning to deliver more in-depth and relevant literature review results. LLMs can propose novel hypotheses by retrieving related literature as inspiration, finding semantically similar content, connecting concepts, and utilising citations based on research topics and motivation[121]. Alternatively, it may require additional ingredients or experiments to extrapolate and search outside current knowledge domains.

This case also highlights the importance of the hypothesis-experiment-observation loop, where each step is critical: hypotheses rely on observations, experiments require hypotheses and planning, and observations depend on experiments. Such a self-dependent loop is typical in scientific discovery and can be initiated either by starting with a tangible step in the hypothesis-experiment-observation process or by allowing human intervention.

**Human Scientists in the Loop**

While we showcase the capabilities of LLMs in assisting scientific discovery, human scientists remain indispensable. During the literature review stage, with the help of LLM agents, humans can contribute by providing deeper perspectives or guiding the focus toward the needs of human scientists[144]. In the reasoning processes, by identifying uncertain reasoning and thoughts, humans can correct LLMs. This significantly improves the accuracy of the chain-of-thought method, making the LLMs more reliable[147]. Human scientists can be involved in further improving safety and reliability. For example,

ORGANA[110], an LLM-powered chemical task robot, uses LLMs to interact with humans via natural language and actively engages with users for disambiguation and troubleshooting. Beyond this, humans can assist LLMs to enhance performance with a reduced workload. For example, by involving humans in the hypothesis-proposing stage to select generated hypotheses, LLMs can perform similarly to humans[133]. At the experiment stage, many lab experiments still require human implementation and correction of invalid experimental plans[67], and LLMs can request human help on these experiments[66].

While the methods described above focus on LLMs as drivers of scientific inquiry, we must clarify that human-in-the-loop is more aptly cast as LLM-in-the-loop, emphasising "assistance" or augmentation as the practical value-added dimension of LLMs. The opportunities described in this paper show potential to shift this mode of scientific practice to be more reliant on AI-driven approaches, but not without significant advances in AIScience approaches with respect to physics-infused ML and causal reasoning and in rigorous testing systems for LLMs interacting with the natural world.

| Scientific Method | Solutions |
| --- | --- |
| Observation | - Replacing human evaluation and annotations [60,61]<br>- Simplifying observed data by providing qualitative descriptions [100,115]<br>- Domain-specific LLMs can perform better on classification and prediction tasks [62-64] |
| Hypotheses | - Literature review: using an LLM's own trained knowledge [142,143,148], or using the RAG method to access up-to-date information [45,92,121].<br>- Novelty: hallucinations of LLMs can sometimes benefit novelty [149]; using the role-play method, LLMs can increase their novelty [136]; LLMs can also propose novel ideas iteratively [121].<br>- Observation-based Hypotheses: LLMs can propose hypotheses based on [100,115,116,139,141]. |
| Experiment | - Implement experiment: LLMs can use external tools [85,96], API calls [66], or directly write code [150] to implement experiments.<br>- Experiment planning: chain-of-thought [44], ReAct [112].<br>- Safety: Hardcoded pipeline for safety [67]; Human confirmation [50] |
| Automation | - LLM agent: LLM-based planner [66], multi-LLM agent [109]<br>- Scaling: complex tasks [116-118]; knowledge accumulation [109]<br>- Enhance: Iteratively optimising proposed hypotheses [121], and experiments [112].<br>- Human-in-the-loop |

Table 1: How LLMs can assist scientific methods at different stages.

## Challenges and Opportunities

While LLMs have shown signs of delivering promising results and of having positive impacts on scientific discovery, investigators have recognised their limitations, such as hallucinations, limited reasoning capabilities, and lack of transparency. Compared to everyday usage, when applied to scientific domains, these limitations require careful consideration, as scientific processes and discoveries require high standards of truthfulness, complex reasoning, and interpretability. The scientific community's increasing recognition and communication of these limitations of LLMs is essential to enabling solutions while also limiting expectations. Such rigour is a cornerstone of science and engineering, and a requirement if LLMs are to play a practical role.

Beyond all this, LLMs also affect scientific research at the scientific community level. While many papers and reviews involve LLMs' assistance, LLMs still face challenges in producing qualified reviews.

### Hallucinations as Putative Sources of Novel Hypotheses

Hallucinations produced by LLMs, also called confabulation or delusion, refer to artificially intelligent systems generating responses that contain false or misleading information presented as fact. This is analogous to hallucination in human psychology, though, for LLMs, it manifests as unjustified responses or beliefs rather than perceptual hallucinations[151]. Hallucinating LLMs can make unsupported claims, thus failing to meet a prior set of standards. While some "incorrect" LLM responses may reflect nuances in the training data not apparent to human reviewers[152], this argument has been challenged as not robust to real-world use[153].

In scientific discovery, hallucination becomes a critical hurdle when applying LLMs to literature review, data processing, or reasoning. Various methods have been developed to mitigate hallucinations[154]. Using the RAG method, LLMs can reference accurate source contexts and up-to-date information, which can reduce hallucinations[155].

Knowledge graphs can also provide reliable information to reduce hallucinations[64]. A self-RAG method can also reduce hallucinations, where the LLMs generate and verify the reference contexts, and outputs are also verified by the LLMs themselves[151]. [156]proposes an even simpler solution: create answers for the same query multiple times and vote for the final answer. This method can significantly improve the accuracy of outputs. Repetition from prompt variation and reiteration can also detect hallucinations–by finding contradictions[157]. By repeatedly generating the same context, LLMs may sometimes generate contradictory content, which can be fixed by the LLMs themselves iteratively.

Another method to mitigate hallucinations is through self-verification. This often involves decomposing the generated content into multiple fact checkpoints. For example, the Chain-of-Verification method uses separate LLM agents to verify them individually and update the original answer[158]. Such a verification process can also adopt RAG methods for greater reliability[159].

An important origin of hallucinations is the auto-regressive generation process of mainstream LLMs, where errors may accumulate during generation[146]. Hence, as discussed above, a general way to mitigate hallucinations is to decompose the end-to-end generation process using chain-of-thoughts, the RAG method, multiple agents, feedback, and iteration loops.

While significant research efforts target the challenge of how to control or limit hallucinations, we may ask to what extent hallucinations are a bug or a feature. For example, could hallucination provide a gateway to creativity in that it could represent a steady stream of novel conjectures? An LLM could then be used to filter such a string of hallucinated hypotheses and rank them to recommend which ones to test. This remains unexplored territory, as far as we can tell.

Another approach to treating hallucinations is to move beyond a binary perspective of trust versus distrust. Instead, similar to statistical confidence, we may quantify the extent to which research conducted by LLM agents can be trusted. Current studies primarily focus on confidence measurements at the foundation model level[160–162] and the output level[163]. Some research has also proposed multidimensional assessments of LLM trustworthiness[164]. Additionally, efforts have been made to enable LLMs to express their confidence levels[160,165]. However, confidence measurements at the LLM agent level are primarily limited to success rates rather than trustworthiness, particularly when dealing with open-ended tasks. Moreover, existing measurements predominantly rely on post-hoc quantifications, which restrict their applicability in scientific research[166]. Therefore, predictive trustworthiness quantification frameworks for LLM agents that collectively consider foundation models, tasks, tools usage, workflow, and external feedback are needed.

**The Value of Reasoning and Interpretation in LLM-led Science**

While LLMs have been suggested to perform reasoning on some tasks, they exhibit severe defects in logical reasoning and serious limitations with respect to common sense reasoning. Notably, while LLMs can correctly answer "X is the capital of Y", they struggle to accurately determine that "Y's capital is X." This is known as the "reversal curse"[167]. Another example is shuffling the order of conditions in a query, which may reduce the performance of LLMs. When the conditions are provided logically, the LLMs can often perform correct inferences but may fail when the conditions do not follow a specific order[168]. LLMs can also fail at simple puzzles, such as determining the odd-numbered dates on a list of famous people's birthdays[169], or in simple questions like "Alice has N brothers, and she also has M sisters. How many sisters does Alice's brother have?" Many LLMs, while achieving high performance on other benchmarks, have shown a lower success rate on this task[170]. When faced with unseen tasks, which are common for human scientists in research, LLMs exhibit a significant drop in accuracy even on simple questions, such as performing 9-base number addition or writing Python code with indexing starting at 1 instead of 0. This suggests that LLMs may rely more on pattern matching than on reasoning, contrary to what many assume[171–173]. Consequently, caution is advised when applying LLMs to novel reasoning tasks, and incorporating human oversight into the process is recommended[173,174]. Another crucial aspect of reasoning is planning capability. As discussed earlier, planning is essential for implementing experiments. While techniques such as ReAct and Reflection demonstrate some planning capabilities in LLMs, their effectiveness remains questionable. Current state-of-the-art LLMs often fail at simple planning tasks[175], such as Blocksword, and are unable to verify the correctness of their plans[175]. In contrast, humans generally excel at creating effective plans for such tasks. However, studies also indicate that integrating LLMs with traditional methods or solvers can enhance planning success rates, reduce research time[175], and provide more flexible ways to interact when developing plans[176].

Some reasoning improvement methods, such as self-correction, can also fail. When LLMs receive feedback without new information or correct labels, self-correction can often lead to compromised performance[177]. Such self-correction prompts may even bias the LLMs, causing them to turn correct answers into incorrect ones. To mitigate this problem, directly including all requirements and criteria in the query prompt is suggested instead of providing them as feedback. This result also indicates that to make corrections, effective feedback needs to include external information, such as experimental results and trustworthy sources[177].

While some progress has been made, more advanced methods are needed to address these reasoning-related challenges. One crucial aspect to consider is consistency – when different LLM agents generate different responses to the same query, the result is considered inconsistent. Notably, the self-consistent method[178] uses LLMs to answer the same question multiple times and chooses the most frequent answer. The answers also help people estimate uncertainty[178], given that LLMs often behave too confidently[179]. Similar methods have also been proposed in[156]. Other methods use different LLM agents to suggest different ideas and then conduct a multi-round debate to arrive at a final answer[180]. As illustrated in Figure 2, these LLM-agent methods can benefit all steps in the hypothesis-experiment-observation loop.

A straightforward but challenging route to scientific discovery is to fine-tune or directly train a model. In[181], the authors propose an innovative solution to the Reversal Curse through "reverse training," which involves training models with both the original and reversed versions of factual statements. Considering the requirements for rigor and prudence in scientific research, attention must be given to the limitations of reasoning tasks.

This is particularly important given that LLMs often exhibit reduced performance in reasoning correctness when encountering novel tasks—a frequent occurrence in scientific research, where the focus is on exploring unknown knowledge.

**The Challenge to Understand LLMs, and the Opportunity to Understand by using LLMs**

A comprehensive scientific interpretation stimulates discussion and further discoveries among scientists. This is especially important for LLM-assisted scientific discovery– given that current LLMs are mostly black boxes, it becomes difficult to trust LLM outputs. To understand LLMs' behaviour, LLMs' language capabilities can be leveraged. There are two types of methods. First, LLMs' hidden states can be used in what are known as probing methods. The Logit Lens method[182] applies the unembedding layer to hidden states or transformed hidden states, enabling semantic understanding of LLMs' hidden states. Representation engineering methods[183] can further detect and control emotion, dishonesty, harmfulness, etc., at the token level, allowing people to read their hidden activities. Besides these, dictionary learning methods can also be used to understand LLMs' hidden states and activations, leading to a fine-grained understanding of LLMs.

The second method is to ask LLMs to explain their reasoning. For instance, the CoT method or reasoning models [56] can explain the thought process before generating results or ask LLMs to explain their reasoning after generating results. However, the self-explanation of LLMs is also questionable. Their explanations are often inconsistent with their behaviours, and we cannot use their explanations to predict their behaviours in counterfactual scenarios[184]. This suggests that LLMs' self-explanation may not accurate and not generalizable. Beyond this, LLMs may also hallucinate in their self-explanations, including content that is not factually grounded[184], making their self-explanations even less trustworthy.

Despite the many difficulties in understanding LLMs, they present a significant opportunity for understanding other systems – they can be used to understand data, interpret other systems, and then prompt humans. By directly showing input and output pairs to LLMs, including language input-output pairs[115], mathematical function input-output pairs, or experimental data, LLMs can be made to explain these black-box systems, including, fMRI data, complex systems like GPT-2, or, potentially, papers written by human scientists that are becoming increasingly difficult to reproduce because of various forces at play (e.g., an increasing number of publications and dubious incentives). This indicates the potential of applying LLMs to explain data and other systems, even though understanding them may still be challenging[116].

This capacity to interpret systems is not limited to human language. Foundation models in specific scientific domains offer domain-grounded interpretability, distinct from the pitfalls of LLM self-explanation. Pre-trained on vast, specialized data, these models learn the "language" inherent in that data. For instance, scGPT in single-cell biology demonstrates this: its learned representations align with established biological knowledge, and it utilizes attention-map visualizations to enhance transparency, elucidating gene interactions that are subsequently validated by domain-specific evidence[76]. Although the faithfulness of attention weight interpretability has been questioned in various cases[185,186], it remains widely used in many AI-for-science applications[187]. Models, including LLMs that use transformer architectures, often inherently benefit from such emergent interpretability for both feature attribution and interaction highlighting.

These approaches support viewing complex domain representations, such as Gene Regulatory Networks (GRNs), as a form of decipherable "domain language." Understanding these intrinsic languages through models whose interpretations are rooted in verifiable domain semantics, rather than potentially unreliable self-explanations, provides a robust method for advancing scientific discovery in specialized fields.

**The Impact on Scientific Practice and the Community**

An open question is how much human science and scientists will be willing to let AI, through technology like LLMs, drive scientific discovery. Would scientists be satisfied letting LLMs set the agenda and conduct experiments with little to no supervision, or do we expect to supervise AI always, driven by the fear of multiple levels of misalignment? This is a misalignment between human scientific interests and the actual practice of science, possibly forcing AI and LLMs to produce data as humans do, with its advantages or disadvantages, including constraining the search space and, therefore, the solution space.

Beyond the individual deployment of the scientific method, scientific discovery also happens at the community level, where scientists publish their work, share ideas, and collaborate. We can consider the scientists and even entire scientific community as an agent that learns from experiments and research publications in a manner similar to reinforcement learning processes[188]. However, learning from failed research (or negative results) is just as important as learning from successful studies[189,190], yet it is currently undervalued[191]. This may be because failed research is far more common than successful research. However, with the massive text-processing capabilities of LLMs, we now have the opportunity to systematically share and learn from failures. Therefore, we advocate for journals and conference to encourage the publication of failed studies and negative results.

This learning process also depend on human values emphasising communication, mutual understanding, and peer review. Evidence shows a significant adoption of LLM-assisted paper writing and peer review in recent years. Estimates indicate that 1% to 10% of papers are written with LLM assistance[192] .In computer science, up to 17.5% of research papers are estimated to be assisted by LLMs, a figure that mainly reflects the output of researchers with strict time constraints[193]. Beyond papers, estimates also show that around 7% $\sim$ 15% of reviews are written with LLM assistance[194,195].

While LLMs can provide feedback that shows a high degree of overlap with human reviewers, they are not proficient at assessing the quality and novelty of research[196] . This limitation is especially significant for high-quality research[197] . Beyond this, LLM-assisted reviews tend to assign higher scores to papers than human reviewers evaluating the same papers[194]. Upon closer examination, LLMs also exhibit a homogenisation problem – they tend to provide similar critiques for different papers[196,198].

Despite LLMs displaying limitations at tasks such as peer reviewing and raising ethical concerns in directly generating academic content, they may still benefit scientific communication. For example, most researchers today are non-native English speakers, so they can benefit from LLMs' language capabilities that fit their diverse demands for proofreading[198], helping alleviate the current bias towards Western Science. On another application, LLMs' code explanation capabilities may help scientists understand poorly documented code, making existing knowledge and work more accessible to a broader range of scientists[198]. With significant growth in using LLMs for writing papers, such impacts will become increasingly important[193].

|  | Challenges | Solution & Opportunities |
|---|---|---|
| Hallucinations | LLMs may generate false or misleading information [[151]] | RAG: external source [[199]], knowledge graph [[200]], self-RAG [[201]]<br>Repeat: sampling and majority vote [[156,157]]<br>Verification: verify the generated content and refine the results [[158,159]] |
| Reasoning | - Reversal curse [[167]]<br>- Order of conditions [[168]]<br>- Alice in Wonderland [[170]]<br>- Limited self-correction [[177]]<br>- Internal reasoning [[169]] | Consistency methods: self-consistency [[178]], multi-agent vote [[156]], multi-agent debate [[180]], Tree-of-Thoughts [[202]]<br>- Learning: Buffer-of-thoughts [[181]]<br>- In-context learning |
| Transparency & Interpretability | - Traditional methods like gradient-based methods are a challenge to apply to LLM due to the scale<br>- Self-explanation is also not trustworthy [[184,203]] | Probing methods: Logit Lens method [[182]], Patchscope [[204]]<br>Interpret data & other systems: explain black-box functions [[100,115]], understand complex neural networks [[116]].<br>Providing symbolic understanding with the help of in-context examples and knowledge [[80]]. |
| Scientific Community | - LLMs are not yet good reviewers [[192,194,195]]<br>- LLMs are widely used in paper and review writing [[192–195]] | LLMs can mitigate the disadvantage of non-native English speakers [[198,205–207]]<br>LLMs can help inter-disciplinary research [[198]] |

Table 2: Challenges and opportunities in applying LLM in scientific discovery. There is much room for AI4Science as a field to fill in gaps under the listed themes, especially as LLM research matures over the next several years; consistent with all innovations in scientific practices and engineering standards, the demonstration-through-validation of said innovations requires time and resources several fold greater than are available in "safer" domains (where LLMs are currently embedded).

**Conclusions**

In this perspective paper, we reviewed the rapid development and integration of large language models (LLMs) in scientific research, highlighting the profound implications of these models for the scientific process. LLMs have evolved from tools of convenience—performing tasks like summarising literature, generating code, and analysing datasets—to emerging as pivotal aids in hypothesis generation, experimental design, and even process automation. As AI advances, foundation models have emerged, representing adaptable, scalable models with the potential to apply across diverse scientific domains, reinforcing the collaborative synergy between humans and machines.

LLMs have reshaped how researchers approach the vast amounts of scientific information available today. By efficiently summarising literature and detecting knowledge gaps, scientists can speed up literature review and idea generation. Furthermore, LLMs facilitate interdisciplinary research, bridging the knowledge divide by summarising complex ideas across fields, thereby fostering collaborations previously limited by domain-specific language and methods. Beyond these benefits, the massive text-processing capabilities of LLMs create new opportunities for utilizing failed research failed research, which has received limited attention. Therefore, we encourage the scientific community to promote the publication of negative results and failed research.

The utility of LLMs in designing experiments is another notable advancement. Models like CRISPR-GPT in biology exemplify this by automating gene-editing experiment designs, significantly accelerating genomics research. Moreover, LLM-powered autonomous systems like BioDiscoveryAgent indicate a shift towards AI-driven experimental processes that can augment researchers' efficiency and, more importantly, enable scientific exploration previously constrained by resource limitations.

So, Large Language Models (LLMs) present two contrasting roles in scientific discovery: accuracy in experimental phases and creativity in hypothesis generation. On the one hand, scientific research requires LLMs to be reliable, accurate, and capable of logical reasoning, particularly for experimental validation. On the other, there is value in promoting creative "hallucinations" or speculative ideas at the hypothesis stage, which mirrors human intuition and expands research boundaries[208].

Besides the general foundation models like GPT-4, Claude and Deepseek, domain specific foundation models have shown special potential for applying LLMs in scientific research Notable examples such as Evo and ChemBERT showcase the success of domain-specific adaptations in genomics and chemistry, where they excel in predicting gene interactions and molecular properties. These foundational models also highlight a promising approach by treating genomic, chemical, and other scientific data as new modalities for LLMs, similar to how images, videos, and audio are considered as modalities. Integrating these modalities often follows two main strategies: end-to-end training, where models like ChemBERT develop deep, intrinsic capabilities on specialized data, potentially exceeding human performance on specific tasks; and compositional approaches, which offer greater flexibility by leveraging intermediate modalities common to human scientists (like vision and text) or specialized tools. While end-to-end methods provide depth, compositional flexibility is crucial for adapting to diverse and rapidly changing scientific demands. Consequently, combining and scaling these scientific modalities, particularly when models can be seamlessly inserted into various scientific workflows, has the potential to profoundly transform scientific research.

Despite the promise, current limitations pose significant hurdles to fully realising LLMs as independent scientific agents. Among these are reasoning limitations, interpretability issues, and challenges like "hallucinations"—where LLMs generate plausible-sounding but inaccurate information. While helpful in generating hypotheses, these models require careful oversight to prevent misleading or unverified information from influencing scientific processes.

The challenges of reasoning and hallucinations pose serious concerns regarding the use of LLMs in scientific discovery. Instead of treating LLMs as simply trustworthy or untrustworthy in a binary manner, we suggest an analogy to statistical confidence, using a continuous value—it may term as *algorithmic confidence*— to quantify the trustworthiness of an LLM agent system in scientific research. We further suggest that all LLM-assisted research should either be verified by humans or undergo algorithmic confidence testing.

The interpretability of LLMs also remains a complex issue. Their black-box nature can obstruct transparency, limiting trust in outputs that affect high-stakes scientific decision-making. Consequently, researchers continue to explore methods such as probing, logit lens techniques, and visualisation of neuron activations to demystify the decision-making processes within these models. Increased interpretability will be critical as we strive for ethically responsible and scientifically sound applications. On the other hand, it is essential to recognize that LLMs are showcasing their potential to explain other black-box systems through their language and reasoning capabilities.

Integrating LLMs into scientific workflows brings ethical considerations, particularly regarding transparency and fairness. For instance, LLMs hold the potential for democratising access to scientific information, aiding researchers from non-English speaking backgrounds in publication and collaborative research. However, they also risk perpetuating biases present in training data, thereby influencing scientific outputs and potentially reinforcing existing disparities in research.

Another concern involves the over-reliance on AI in scientific processes. As we incorporate LLMs deeper into workflows, human oversight becomes essential to maintaining scientific rigor and addressing potential misalignments between AI-generated outputs and human-defined research goals. The question of how much autonomy AI should have in guiding scientific inquiries raises ongoing debate about accountability and the evolving role of human oversight.

To harness LLMs as creativity engines, moving beyond task-oriented applications to generate new scientific hypotheses and theories is paramount. For LLMs to contribute meaningfully to fundamental scientific discoveries, they must be equipped to recognize patterns and autonomously generate novel, insightful questions—a hallmark of scientific creativity. This would require advancements in prompt engineering, automated experimentation, iterative reasoning, and building an AI that evolves its approach based on experimental feedback. However, a significant gap in general reasoning capabilities separates current models from domain-specific superhuman systems like AlphaGo/AlphaZero[209,210]. AlphaGo leveraged a critical symmetry where an "answer" (a move) inherently generates a new "question" (the next board state challenge)—a dynamic largely absent in today's reasoning models, yet key for mastering novel tasks. For scientific discovery, developing this symmetry is crucial, as the ability to ask questions is as important as answering them; though some preliminary work has explored this[211], it remains an unsolved and highly challenging problem.

The evolution of LLMs and foundation models signals a transformative era for science. While current applications largely support scientists in managing data and expediting workflows, the future may see these models as integral components of the scientific process. By addressing challenges in accuracy, interpretability, and ethical concerns, we can enhance their reliability and pave the way for responsible AI in scientific contexts.

Looking ahead, the collaboration between AI and human scientists will likely define the next generation of discovery. As we refine foundation models to become more adaptable and creative, they may transition from merely assisting to potentially leading explorations into uncharted scientific domains. The challenge lies in responsibly developing these models to ensure they complement and elevate human expertise without compromising scientific integrity. Ultimately, LLMs and foundation models may come to represent a synthesis of human and artificial intelligence, each amplifying the strengths of the other. With continued research and ethical vigilance, LLMs have the potential to accelerate and deepen scientific discovery, heralding a new era where AI not only supports but inspires new frontiers in science[212]. This includes embracing and learning from scientific failures and leveraging them to drive comprehensive exploration. As LLMs evolve, they may reshape scientific methodologies, impacting how science values discovery and reproducibility and may ultimately redefine the purpose of scientific inquiry.

## References

1. Wang, H. *et al.* Scientific discovery in the age of artificial intelligence. *Nature* **620**, 47–60 (2023).
2. Alkhateeb, A. & Aeon. Can Scientific Discovery Be Automated? *The Atlanctic* (2017).
3. Jain, M. *et al.* GFlowNets for AI-driven scientific discovery. *Digital Discovery* **2**, 557–577 (2023).
4. Kitano, H. Nobel Turing Challenge: creating the engine for scientific discovery. *NPJ Syst Biol Appl* **7**, 29 (2021).
5. Cornelio, C. *et al.* Combining data and theory for derivable scientific discovery with AI-Descartes. *Nature Communications 2023 14:1* **14**, 1–10 (2023).
6. Kim, S. *et al.* Integration of Neural Network-Based Symbolic Regression in Deep Learning for Scientific Discovery. *IEEE Trans Neural Netw Learn Syst* **32**, 4166–4177 (2021).
7. Gil, Y., Greaves, M., Hendler, J. & Hirsh, H. Amplify scientific discovery with artificial intelligence: Many human activities are a bottleneck in progress. *Science (1979)* **346**, 171–172 (2014).
8. Kitano, H. Artificial Intelligence to Win the Nobel Prize and Beyond: Creating the Engine for Scientific Discovery. *AI Mag* **37**, 39–49 (2016).
9. Berens, P., Cranmer, K., Lawrence, N. D., von Luxburg, U. & Montgomery, J. AI for Science: An Emerging Agenda. (2023).

10. Li, Z., Ji, J. & Zhang, Y. *From Kepler to Newton: Explainable AI for Science Discovery*. (2022).
11. Baker, N. *et al.* Workshop Report on Basic Research Needs for Scientific Machine Learning: Core Technologies for Artificial Intelligence. (2019) doi:10.2172/1478744.
12. Manta, C. D., Hu, E. & Bengio, Y. GFlowNets for Causal Discovery: an Overview. *OpenReview* (2023).
13. Vinuesa, R., Brunton, S. L. & McKeon, B. J. The transformative potential of machine learning for experiments in fluid mechanics. *Nature Reviews Physics* **5**, 536–545 (2023).
14. del Rosario, Z. & del Rosario, M. Synthesizing domain science with machine learning. *Nat Comput Sci* **2**, 779–780 (2022).
15. Krenn, M., Landgraf, J., Foesel, T. & Marquardt, F. Artificial Intelligence and Machine Learning for Quantum Technologies. (2022).
16. van der Schaar, M. *et al.* How artificial intelligence and machine learning can help healthcare systems respond to COVID-19. *Mach Learn* **110**, 1–14 (2021).
17. Zhang, T. *et al.* AI for Global Climate Cooperation: Modeling Global Climate Negotiations, Agreements, and Long-Term Cooperation in RICE-N. (2022).
18. Pion-Tonachini, L. *et al.* Learning from learning machines: a new generation of AI technology to meet the needs of science. (2021).
19. Keith, J. A. *et al.* Combining Machine Learning and Computational Chemistry for Predictive Insights into Chemical Systems. *Chemical Reviews* vol. 121 9816–9872 Preprint at https://doi.org/10.1021/acs.chemrev.1c00107 (2021).
20. Birhane, A., Kasirzadeh, A., Leslie, D. & Wachter, S. Science in the age of large language models. *Nature Reviews Physics* **5**, 277–280 (2023).
21. Georgescu, I. How machines could teach physicists new scientific concepts. *Nature Reviews Physics* **4**, 736–738 (2022).
22. Noé, F., Tkatchenko, A., Müller, K.-R. & Clementi, C. Machine Learning for Molecular Simulation. *Annu Rev Phys Chem* (2020) doi:10.1146/annurev-physchem-042018.
23. Moor, M. *et al.* Foundation models for generalist medical artificial intelligence. *Nature* **616**, 259–265 (2023).
24. Rajpurkar, P., Chen, E., Banerjee, O. & Topol, E. J. AI in health and medicine. *Nat Med* **28**, 31–38 (2022).
25. Acosta, J. N., Falcone, G. J., Rajpurkar, P. & Topol, E. J. Multimodal biomedical AI. *Nat Med* (2022) doi:10.1038/s41591-022-01981-2.
26. Topol, E. J. Welcoming new guidelines for AI clinical research. *Nat Med* **26**, 1318–1320 (2020).
27. Topol, E. Deep-Medicine. (2019).
28. Krishnan, R., Rajpurkar, P. & Topol, E. J. Self-supervised learning in medicine and healthcare. *Nat Biomed Eng* (2022) doi:10.1038/s41551-022-00914-1.
29. Stoyanovich, J., Bavel, J. J. Van & West, T. V. The imperative of interpretable machines. *Nat Mach Intell* **2**, 197–199 (2020).
30. Meskó, B. & Topol, E. J. The imperative for regulatory oversight of large language models (or generative AI) in healthcare. *NPJ Digit Med* **6**, (2023).

31. Willcox, K. E., Ghattas, O. & Heimbach, P. The imperative of physics-based modeling and inverse theory in computational science. *Nat Comput Sci* **1**, 166–168 (2021).
32. Webster, P. Six ways large language models are changing healthcare. *Nat Med* **29**, 2969–2971 (2023).
33. Eriksen, A. V, Möller, S. & Ryg, J. Use of GPT-4 to Diagnose Complex Clinical Cases. *NEJM AI* **1**, (2023).
34. Ishmam, F., Sakib, M., Shovon, H., Mridha, M. F. & Dey, N. From Image to Language: A Critical Analysis of Visual Question Answering (VQA) Approaches, Challenges, and Opportunities. Preprint at https://www.bemyeyes.com/ (2023).
35. Li, C. *et al.* Multimodal Foundation Models: From Specialists to General-Purpose Assistants. (2023).
36. Omiye, J. A., Gui, H., Rezaei, S. J., Zou, J. & Daneshjou, R. Large language models in medicine: the potentials and pitfalls. Preprint at (2023).
37. Maithra Raghu; Eric Schmidt. A Survey of Deep Learning for Scientific Discovery. *arXiv preprint arXiv:2003.11755* (2020).
38. Song, S. *et al.* DeepSpeed4Science Initiative: Enabling Large-Scale Scientific Discovery through Sophisticated AI System Technologies. in *NeurIPS 2023 AI for Science Workshop* (2023).
39. McCarthy, J. , M. M. , R. N. , & S. C. E. Dartmouth Summer Research Project on Artificial Intelligence. in *Dartmouth Summer Research Project on Artificial Intelligence* (1956).
40. Ramos, M. C., Collison, C. J. & White, A. D. A Review of Large Language Models and Autonomous Agents in Chemistry. (2024).
41. Zenil, H. & King, R. Artificial intelligence in scientific discovery: Challenges and opportunities. in *Science: Challenges, Opportunities and the Future of Research* ( OECD Publishing, Paris, 2023).
42. Zenil, H. & King, R. A framework for evaluating the AI-driven automation of science. in *Science: Challenges, Opportunities and the Future of Research* (OECD Publishing, Paris, 2023).
43. Zenil, H. & King, R. The Far Future of AI in Scientific Discovery. in *AI For Science* (ed. Choudhary, F. and T. H.) (World Scientific, 2023).
44. Wei, J. *et al.* Chain-of-Thought Prompting Elicits Reasoning in Large Language Models. *Adv Neural Inf Process Syst* **35**, (2022).
45. Lewis, P. *et al.* Retrieval-augmented generation for knowledge-intensive nlp tasks. *Adv Neural Inf Process Syst* **33**, 9459–9474 (2020).
46. White, J. *et al.* A Prompt Pattern Catalog to Enhance Prompt Engineering with ChatGPT. (2023).
47. Schulhoff, S. *et al.* The Prompt Report: A Systematic Survey of Prompting Techniques. (2024).
48. Fulford, I. & Ng, A. ChatGPT Prompt Engineering for Developers. (2024).
49. Cheng, Y. *et al.* Exploring large language model based intelligent agents: Definitions,    methods, and prospects. *arXiv preprint arXiv:2401. 03428* (2024).
50. Yang, H., Yue, S. & He, Y. Auto-GPT for Online Decision Making: Benchmarks and Additional Opinions. (2023).

51. Nakajima, Y. yoheinakajima/babyagi. Preprint at https://github.com/yoheinakajima/babyagi (2024).
52. Chase, H. LangChain. https://github.com/langchain-ai/langchain (2022).
53. Liu, J. LlamaIndex. *URL: 'https://github.com/jerryjliu/llama_index'* (2022).
54. Khattab, O. *et al.* DSPy: Compiling Declarative Language Model Calls into Self-Improving Pipelines. *12th International Conference on Learning Representations, ICLR 2024* (2023).
55. Yuksekgonul, M. *et al.* TextGrad: Automatic 'Differentiation' via Text. (2024).
56. DeepSeek-AI *et al.* DeepSeek-R1: Incentivizing Reasoning Capability in LLMs via Reinforcement Learning. (2025).
57. Shao, Z. *et al.* DeepSeekMath: Pushing the Limits of Mathematical Reasoning in Open Language Models. (2024).
58. Li, Z.-Z. *et al.* From System 1 to System 2: A Survey of Reasoning Large Language Models. (2025).
59. Caufield, J. H. *et al.* Structured prompt interrogation and recursive extraction of semantics (SPIRES): A method for populating knowledge bases using zero-shot learning. *Bioinformatics* **40**, btae104 (2024).
60. Gilardi, F., Alizadeh, M. & Kubli, M. ChatGPT outperforms crowd workers for text-annotation tasks. *Proceedings of the National Academy of Sciences* **120**, e2305016120 (2023).
61. Liu, Y. *et al.* G-Eval: NLG Evaluation using Gpt-4 with Better Human Alignment. *EMNLP 2023 - 2023 Conference on Empirical Methods in Natural Language Processing, Proceedings* 2511–2522 (2023) doi:10.18653/V1/2023.EMNLP-MAIN.153.
62. Gu, Y. *et al.* Domain-specific language model pretraining for biomedical natural language processing. *ACM Transactions on Computing for Healthcare (HEALTH)* **3**, 1–23 (2021).
63. Irwin, R., Dimitriadis, S., He, J. & Bjerrum, E. J. Chemformer: a pre-trained transformer for computational chemistry. *Mach Learn Sci Technol* **3**, 015022 (2022).
64. Luo, R. *et al.* BioGPT: generative pre-trained transformer for biomedical text generation and mining. *Brief Bioinform* **23**, bbac409 (2022).
65. Gottweis, J. *et al.* Towards an AI co-scientist. (2025).
66. Boiko, D. A., MacKnight, R. & Gomes, G. Emergent autonomous scientific research capabilities of large language models. *arXiv preprint arXiv:2304. 05332* (2023).
67. M. Bran, A. *et al.* Augmenting large language models with chemistry tools. *Nat Mach Intell* 1–11 (2024).
68. Qu, Y. *et al. CRISPR-GPT: An LLM Agent for Automated Design of Gene-Editing Experiments*. *bioRxiv* https://www.biorxiv.org/content/10.1101/2024.04.25.591003v2 (2024) doi:10.1101/2024.04.25.591003.
69. Roohani, Y. *et al.* BioDiscoveryAgent: An AI Agent for Designing Genetic Perturbation Experiments. *arXiv preprint arXiv:2405. 17631* (2024).
70. OpenAI *et al.* OpenAI o1 System Card. *arXiv:2305.14947v2* (2024).

71. OpenAI. Learning to Reason with LLMs. https://openai.com/index/learning-to-reason-with-llms/ (2024).
72. Kaplan, J. *et al.* Scaling laws for neural language models. *arXiv preprint arXiv:2001. 08361* (2020).
73. Wei, J. *et al.* Emergent abilities of large language models. *arXiv preprint arXiv:2206. 07682* (2022).
74. Nguyen, E. *et al.* Sequence modeling and design from molecular to genome scale with Evo. *Science (1979)* **386**, (2024).
75. Brixi, G. *et al.* Genome modeling and design across all domains of life with Evo 2. *bioRxiv* 2025.02.18.638918 (2025) doi:10.1101/2025.02.18.638918.
76. Cui, H. *et al.* scGPT: toward building a foundation model for single-cell multi-omics using generative AI. *Nature Methods 2024 21:8* **21**, 1470–1480 (2024).
77. Chithrananda, S., Grand, G. & Ramsundar, B. ChemBERTa: Large-Scale Self-Supervised Pretraining for Molecular Property Prediction. (2020).
78. Birk, J., Hallin, A. & Kasieczka, G. OmniJet-$\alpha$: The first cross-task foundation model for particle physics. *Mach Learn Sci Technol* **5**, (2024).
79. McCabe, M. *et al.* Multiple Physics Pretraining for Physical Surrogate Models. (2023).
80. Shojaee, P., Meidani, K., Gupta, S., Farimani, A. B. & Reddy, C. K. LLM-SR: Scientific Equation Discovery via Programming with Large Language Models. *ArXiv* **abs/2404.18400**, (2024).
81. Zhang, S. *et al.* BiomedCLIP: a multimodal biomedical foundation model pretrained from fifteen million scientific image-text pairs. (2023).
82. Mishra-Sharma, S., Song, Y. & Thaler, J. PAPERCLIP: Associating Astronomical Observations and Natural Language with Multi-Modal Models. (2024).
83. Parker, L. *et al.* AstroCLIP: a cross-modal foundation model for galaxies. *Mon Not R Astron Soc* **531**, 4990–5011 (2024).
84. M. Bran, A. *et al.* Augmenting large language models with chemistry tools. *Nature Machine Intelligence 2024 6:5* **6**, 525–535 (2024).
85. OpenAI *et al.* GPT-4 Technical Report. (2023).
86. Devlin, J., Chang, M. W., Lee, K. & Toutanova, K. BERT: Pre-training of Deep Bidirectional Transformers for Language Understanding. *NAACL HLT 2019 - 2019 Conference of the North American Chapter of the Association for Computational Linguistics: Human Language Technologies - Proceedings of the Conference* **1**, 4171–4186 (2018).
87. Radford, A. *et al.* Learning Transferable Visual Models From Natural Language Supervision. *Proc Mach Learn Res* **139**, 8748–8763 (2021).
88. Ramesh, A. *et al.* Zero-Shot Text-to-Image Generation. *Proc Mach Learn Res* **139**, 8821–8831 (2021).
89. DeepMind. AlphaProof : AI achieves silver-medal standard solving International Mathematical Olympiad problems. https://deepmind.google/discover/blog/ai-solves-imo-problems-at-silver-medal-level/ (2024).
90. Zelikman, E., Wu, Y., Mu, J. & Goodman, N. D. STaR: Bootstrapping Reasoning With Reasoning. (2022).
91. Popper, K. *Karl Popper: The Logic of Scientific Discovery*. (1959).

92. Fan, W. *et al.* A Survey on RAG Meeting LLMs: Towards Retrieval-Augmented Large Language Models. *Proceedings of the ACM SIGKDD International Conference on Knowledge Discovery and Data Mining* 6491–6501 (2024) doi:10.1145/3637528.3671470.
93. Li, Y., Xu, M., Miao, X., Zhou, S. & Qian, T. Prompting Large Language Models for Counterfactual Generation: An Empirical Study. *2024 Joint International Conference on Computational Linguistics, Language Resources and Evaluation, LREC-COLING 2024 - Main Conference Proceedings* 13201–13221 (2023).
94. Kıcıman, E., Ness, R., Sharma, A. & Tan, C. Causal reasoning and large language models: Opening a new frontier for causality. *arXiv preprint arXiv:2305. 00050* (2023).
95. Jiralerspong, T., Chen, X., More, Y., Shah, V. & Bengio, Y. Efficient Causal Graph Discovery Using Large Language Models. *arXiv preprint arXiv:2402. 01207* (2024).
96. Schick, T. *et al.* Toolformer: Language Models Can Teach Themselves to Use Tools. *Adv Neural Inf Process Syst* **36**, (2023).
97. Chen, Q., Ho, Y.-J. I., Sun, P. & Wang, D. The Philosopher's Stone for Science--The Catalyst Change of AI for Scientific Creativity. *Pin and Wang, Dashun, The Philosopher's Stone for Science--The Catalyst Change of AI for Scientific Creativity (March 5, 2024)* (2024).
98. Zenil, H. *et al.* An Algorithmic Information Calculus for Causal Discovery and Reprogramming Systems. *iScience* **19**, 1160–1172 (2019).
99. Pattee, H. H., Rączaszek-Leonardi, J. & Pattee, H. H. Evolving self-reference: matter, symbols, and semantic closure. *Laws, Language and Life: Howard Pattee's classic papers on the physics of symbols with contemporary commentary* 211–226 (2012).
100. Singh, C., Morris, J. X., Aneja, J., Rush, A. M. & Gao, J. Explaining patterns in data with language models via interpretable autoprompting. *arXiv preprint arXiv:2210. 01848* (2022).
101. Li, P. *et al.* Table-gpt: Table-tuned gpt for diverse table tasks. *arXiv preprint arXiv:2310. 09263* (2023).
102. Zhan, J. *et al.* Anygpt: Unified multimodal llm with discrete sequence modeling. *arXiv preprint arXiv:2402. 12226* (2024).
103. Wu, S. *et al.* Beyond Language Models: Byte Models are Digital World Simulators. (2024).
104. Radford, A. *et al.* Language models are unsupervised multitask learners. *OpenAI blog* **1**, 9 (2019).
105. Weininger, D. SMILES, a chemical language and information system. 1. Introduction to methodology and encoding rules. *J Chem Inf Comput Sci* **28**, 31–36 (1988).
106. Zhang, S. *et al.* Intelligence at the Edge of Chaos. (2024).
107. Lyu, C. *et al.* Large Language Models as Code Executors: An Exploratory Study. (2024).
108. Jiang, J., Wang, F., Shen, J., Kim, S. & Kim, S. A Survey on Large Language Models for Code Generation. (2024).

109. Wang, G. *et al.* Voyager: An open-ended embodied agent with large language models. *arXiv preprint arXiv:2305. 16291* (2023).
110. Darvish, K. *et al.* ORGANA: A Robotic Assistant for Automated Chemistry Experimentation and Characterization. (2024).
111. Shinn, N., Cassano, F., Gopinath, A., Narasimhan, K. & Yao, S. Reflexion: Language Agents with Verbal Reinforcement Learning. *Adv Neural Inf Process Syst* **36**, (2023).
112. Yao, S. *et al.* React: Synergizing reasoning and acting in language models. *arXiv preprint arXiv:2210. 03629* (2022).
113. Lavin, A. *et al.* Simulation Intelligence: Towards a New Generation of Scientific Methods. *ArXiv* **abs/2112.03235**, (2021).
114. Park, J. S. *et al.* Generative Agents: Interactive Simulacra of Human Behavior. *UIST 2023 - Proceedings of the 36th Annual ACM Symposium on User Interface Software and Technology* (2023) doi:10.1145/3586183.3606763.
115. Singh, C. *et al.* Explaining black box text modules in natural language with language models. *arXiv preprint arXiv:2305. 09863* (2023).
116. Bills, S. *et al.* Language models can explain neurons in language models. *URL https://openaipublic. blob. core. windows. net/neuron-explainer/paper/index. html. (Date accessed: 14. 05. 2023)* (2023).
117. Yin, Y., Wang, Y., Evans, J. A. & Wang, D. Quantifying the dynamics of failure across science, startups and security. *Nature* **575**, 190–194 (2019).
118. Kauffman, S. A. *Investigations*. (Oxford University Press, 2000).
119. Burger, B. *et al.* A mobile robotic chemist. *Nature 2020 583:7815* **583**, 237–241 (2020).
120. Yang, C. *et al.* Large language models as optimizers. *arXiv preprint arXiv:2309. 03409* (2023).
121. Wang, Q., Downey, D., Ji, H. & Hope, T. SciMON: Scientific Inspiration Machines Optimized for Novelty. *arXiv preprint arXiv:2305. 14259* (2023).
122. Ellis, K. *et al.* DreamCoder: growing generalizable, interpretable knowledge with wake–sleep Bayesian program learning. *Philosophical Transactions of the Royal Society A* **381**, (2020).
123. Hazan, H. & Levin, M. Exploring The Behavior of Bioelectric Circuits using Evolution Heuristic Search. *Bioelectricity* **4**, 207–227 (2022).
124. Fleming, L. Recombinant uncertainty in technological search. *Manage Sci* **47**, 117–132 (2001).
125. Weitzman, M. *Optimal Search for the Best Alternative*. vol. 78 (Department of Energy, 1978).
126. Van Dis, E. A. M., Bollen, J., Zuidema, W., van Rooij, R. & Bockting, C. L. ChatGPT: five priorities for research. *Nature* **614**, 224–226 (2023).
127. Edwards, B. Why ChatGPT and Bing Chat are so good at making things up. *Ars Technica* (2023).
128. DeepMind. Shaking the foundations: delusions in sequence models for interaction and control. *www. deepmind. com* (2023).
129. Wang, R. *et al.* Hypothesis search: Inductive reasoning with language models. *arXiv preprint arXiv:2309. 05660* (2023).

130. Lavin, A. *et al.* Technology readiness levels for machine learning systems. *Nat Commun* **13**, (2020).
131. Zhou, J. P. *et al.* Don't Trust: Verify -- Grounding LLM Quantitative Reasoning with Autoformalization. *12th International Conference on Learning Representations, ICLR 2024* (2024).
132. Olausson, T. X. *et al.* LINC: A Neurosymbolic Approach for Logical Reasoning by Combining Language Models with First-Order Logic Provers. *EMNLP 2023 - 2023 Conference on Empirical Methods in Natural Language Processing, Proceedings* 5153–5176 (2023) doi:10.18653/V1/2023.EMNLP-MAIN.313.
133. Wang, R. *et al.* Hypothesis Search: Inductive Reasoning with Language Models. *12th International Conference on Learning Representations, ICLR 2024* (2023).
134. Chollet, F. On the Measure of Intelligence. (2019).
135. Abdel-Rehim, A. *et al.* Scientific Hypothesis Generation by a Large Language Model: Laboratory Validation in Breast Cancer Treatment. Royal Society Interface;22: 20240674 (2025).
136. Zhao, Y. *et al.* Assessing and Understanding Creativity in Large Language Models. arXiv:2401.12491 [cs.CL] (2024).
137. Swanson, K., Wu, W., Bulaong, N. L., Pak, J. E. & Zou, J. The Virtual Lab: AI Agents Design New SARS-CoV-2 Nanobodies with Experimental Validation. *bioRxiv* 2024.11.11.623004 (2024) doi:10.1101/2024.11.11.623004.
138. Girotra, K., Meincke, L., Terwiesch, C. & Ulrich, K. T. Ideas are dimes a dozen: Large language models for idea generation in innovation. *Available at SSRN 4526071* (2023).
139. Qi, B. *et al.* Large Language Models are Zero Shot Hypothesis Proposers. *arXiv preprint arXiv:2311. 05965* (2023).
140. Singh, C. *et al.* Explaining black box text modules in natural language with language models. (2023).
141. Liu, T. J. B., Boullé, N., Sarfati, R. & Earls, C. J. LLMs learn governing principles of dynamical systems, revealing an. *arXiv preprint arXiv:2402. 00795* (2024).
142. Delétang, G. *et al.* Language Modeling Is Compression. *12th International Conference on Learning Representations, ICLR 2024* (2023).
143. Huang, Y., Zhang, J., Shan, Z. & He, J. Compression Represents Intelligence Linearly. (2024).
144. Shao, Y. *et al.* Assisting in Writing Wikipedia-like Articles From Scratch with Large Language Models. *Proceedings of the 2024 Conference of the North American Chapter of the Association for Computational Linguistics: Human Language Technologies, NAACL 2024* **1**, 6252–6278 (2024).
145. OpenAI. Introducing Deep Research. https://openai.com/index/introducing-deep-research/ (2025).
146. Perplexity AI. Introducing Perplexity Deep Research. https://www.perplexity.ai/hub/blog/introducing-perplexity-deep-research (2025).
147. Cai, Z., Chang, B. & Han, W. Human-in-the-Loop through Chain-of-Thought. (2023).
148. Petroni, F. *et al.* Language Models as Knowledge Bases? *EMNLP-IJCNLP 2019 - 2019 Conference on Empirical Methods in Natural Language Processing and 9th*

*International Joint Conference on Natural Language Processing, Proceedings of the Conference* 2463–2473 (2019) doi:10.18653/v1/d19-1250.
149. Lee, M. A mathematical investigation of hallucination and creativity in gpt models. *Mathematics* **11**, 2320 (2023).
150. Chen, M. *et al.* Evaluating Large Language Models Trained on Code. (2021).
151. Ji, Z. *et al.* Survey of Hallucination in Natural Language Generation. *ACM Comput Surv* **55**, 1–38 (2022).
152. Matsakis, L. Artificial Intelligence May Not 'Hallucinate' After All. *Wired* (2019).
153. Gilmer, J. & Hendrycks, D. A Discussion of 'Adversarial Examples Are Not Bugs, They Are Features':     Adversarial Example Researchers Need to Expand What is Meant by 'Robustness'. *Distill* **4**, 00019.1 (2019).
154. Tonmoy, S. M. T. I. *et al.* A Comprehensive Survey of Hallucination Mitigation Techniques in Large Language Models. (2024).
155. Varshney, N., Yao, W., Zhang, H., Chen, J. & Yu, D. A Stitch in Time Saves Nine: Detecting and Mitigating Hallucinations of LLMs by Validating Low-Confidence Generation. (2023).
156. Li, J., Zhang, Q., Yu, Y., Fu, Q. & Ye, D. More Agents Is All You Need. (2024).
157. Mündler, N., He, J., Jenko, S. & Vechev, M. Self-contradictory Hallucinations of Large Language Models: Evaluation, Detection and Mitigation. (2023).
158. Dhuliawala, S. *et al.* Chain-of-Verification Reduces Hallucination in Large Language Models. *Findings of the Association for Computational Linguistics ACL 2024* 3563–3578 (2024) doi:10.18653/V1/2024.FINDINGS-ACL.212.
159. Gao, L. *et al.* RARR: Researching and Revising What Language Models Say, Using Language Models. *Proceedings of the Annual Meeting of the Association for Computational Linguistics* **1**, 16477–16508 (2023).
160. Zhu, C., Xu, B., Wang, Q., Zhang, Y. & Mao, Z. On the Calibration of Large Language Models and Alignment. (2022).
161. Li, J., Cheng, X., Zhao, W. X., Nie, J. Y. & Wen, J. R. HaluEval: A Large-Scale Hallucination Evaluation Benchmark for Large Language Models. *EMNLP 2023 - 2023 Conference on Empirical Methods in Natural Language Processing, Proceedings* 6449–6464 (2023) doi:10.18653/V1/2023.EMNLP-MAIN.397.
162. Lin, S., Hilton, J. & Evans, O. TruthfulQA: Measuring How Models Mimic Human Falsehoods. *Proceedings of the Annual Meeting of the Association for Computational Linguistics* **1**, 3214–3252 (2021).
163. Min, S. *et al.* FActScore: Fine-grained Atomic Evaluation of Factual Precision in Long Form Text Generation. *EMNLP 2023 - 2023 Conference on Empirical Methods in Natural Language Processing, Proceedings* 12076–12100 (2023) doi:10.18653/v1/2023.emnlp-main.741.
164. Liu, Y. *et al.* Trustworthy LLMs: a Survey and Guideline for Evaluating Large Language Models' Alignment. (2023).
165. Li, M. *et al.* Think Twice Before Trusting: Self-Detection for Large Language Models through Comprehensive Answer Reflection. (2024).
166. Ye, Q., Fu, H. Y., Ren, X. & Jia, R. How Predictable Are Large Language Model Capabilities? A Case Study on BIG-bench. *Findings of the Association for Computational Linguistics: EMNLP 2023* 7493–7517 (2023) doi:10.18653/v1/2023.findings-emnlp.503.

167. Berglund, L. *et al.* The Reversal Curse: LLMs trained on 'A is B' fail to learn 'B is A'. *arXiv preprint arXiv:2309. 12288* (2023).
168. Chen, X., Chi, R. A., Wang, X. & Zhou, D. Premise Order Matters in Reasoning with Large Language Models. *Proc Mach Learn Res* **235**, 6596–6620 (2024).
169. Allen-Zhu, Z. & Labs Yuanzhi Li YuanzhiLi, F. Physics of Language Models: Part 3.2, Knowledge Manipulation. (2023).
170. Nezhurina, M., Cipolina-Kun, L., Cherti, M. & Jitsev, J. Alice in Wonderland: Simple Tasks Showing Complete Reasoning Breakdown in State-Of-the-Art Large Language Models. (2024).
171. Dziri, N. *et al.* Faith and Fate: Limits of Transformers on Compositionality. *Adv Neural Inf Process Syst* **36**, (2023).
172. Delétang, G. *et al.* Neural Networks and the Chomsky Hierarchy. *11th International Conference on Learning Representations, ICLR 2023* (2022).
173. McCoy, R. T., Yao, S., Friedman, D., Hardy, M. D. & Griffiths, T. L. Embers of autoregression show how large language models are shaped by the problem they are trained to solve. *Proc Natl Acad Sci U S A* **121**, e2322420121 (2024).
174. Wu, Z. *et al.* Reasoning or Reciting? Exploring the Capabilities and Limitations of Language Models Through Counterfactual Tasks. *Proceedings of the 2024 Conference of the North American Chapter of the Association for Computational Linguistics: Human Language Technologies, NAACL 2024* **1**, 1819–1862 (2024).
175. Kambhampati, S. *et al.* LLMs Can't Plan, But Can Help Planning in LLM-Modulo Frameworks. (2024).
176. Pallagani, V. *et al.* On the Prospects of Incorporating Large Language Models (LLMs) in Automated Planning and Scheduling (APS). *Proceedings of the International Conference on Automated Planning and Scheduling* **34**, 432–444 (2024).
177. Huang, J. *et al.* Large Language Models Cannot Self-Correct Reasoning Yet. *12th International Conference on Learning Representations, ICLR 2024* (2023).
178. Wang, X. *et al.* Self-Consistency Improves Chain of Thought Reasoning in Language Models. *11th International Conference on Learning Representations, ICLR 2023* (2022).
179. Zhou, K., Hwang, J. D., Ren, X. & Sap, M. Relying on the Unreliable: The Impact of Language Models' Reluctance to Express Uncertainty. *arXiv preprint arXiv:2401. 06730* (2024).
180. Du, Y., Li, S., Torralba, A., Tenenbaum, J. B. & Mordatch, I. Improving Factuality and Reasoning in Language Models through Multiagent Debate. *Proc Mach Learn Res* **235**, 11733–11763 (2023).
181. Golovneva, O., Allen-Zhu, Z., Weston, J. & Sukhbaatar, S. Reverse Training to Nurse the Reversal Curse. *arXiv preprint arXiv:2403. 13799* (2024).
182. interpreting GPT: the logit lens — LessWrong. https://www.lesswrong.com/posts/AcKRB8wDpdaN6v6ru/interpreting-gpt-the-logit-lens.
183. Zou, A. *et al.* Representation engineering: A top-down approach to ai transparency. *arXiv preprint arXiv:2310. 01405* (2023).
184. Chen, Y. *et al.* Do models explain themselves? Counterfactual simulatability of natural language explanations. *arXiv preprint arXiv:2307. 08678* (2023).

185. Wiegreffe, S. & Pinter, Y. Attention is not not Explanation. *EMNLP-IJCNLP 2019 - 2019 Conference on Empirical Methods in Natural Language Processing and 9th International Joint Conference on Natural Language Processing, Proceedings of the Conference* 11–20 (2019) doi:10.18653/V1/D19-1002.
186. Jain, S. & Wallace, B. C. Attention is not Explanation. *Proceedings of the 2019 Conference of the North* 3543–3556 (2019) doi:10.18653/V1/N19-1357.
187. Zhang, Y. *et al.* Attention is all you need: utilizing attention in AI-enabled drug discovery. *Brief Bioinform* **25**, 1–22 (2023).
188. Narayanan, S. *et al.* Aviary: training language agents on challenging scientific tasks. (2024).
189. Taragin, M. I. Learning from negative findings. *Isr J Health Policy Res* **8**, 1–4 (2019).
190. Bik, E. M. Publishing negative results is good for science. *Access Microbiol* **6**, 000792 (2024).
191. Echevarriá, L., Malerba, A. & Arechavala-Gomeza, V. Researcher's Perceptions on Publishing "Negative" Results and Open Access. *Nucleic Acid Ther* **31**, 185 (2021).
192. Gray, A. ChatGPT 'contamination': estimating the prevalence of LLMs in the scholarly literature. (2024) doi:10.1002/leap.1578.
193. Liang, W. *et al.* Mapping the Increasing Use of LLMs in Scientific Papers. (2024).
194. Latona, G. R., Ribeiro, M. H., Davidson, T. R., Veselovsky, V. & West, R. The AI Review Lottery: Widespread AI-Assisted Peer Reviews Boost Paper Scores and Acceptance Rates. (2024).
195. Liang, W. *et al.* Monitoring AI-Modified Content at Scale: A Case Study on the Impact of ChatGPT on AI Conference Peer Reviews. 29575–29620 Preprint at https://proceedings.mlr.press/v235/liang24b.html (2024).
196. Liang, W. *et al.* Can Large Language Models Provide Useful Feedback on Research Papers? A Large-Scale Empirical Analysis. *NEJM AI* **1**, (2024).
197. Thelwall, M. Can ChatGPT evaluate research quality? *Journal of Data and Information Science* **9**, 1–21 (2024).
198. Meyer, J. G. *et al.* ChatGPT and large language models in academia: opportunities and challenges. *BioData Min* **16**, 20 (2023).
199. Peng, B. *et al.* Check your facts and try again: Improving large language models with external knowledge and automated feedback. *arXiv preprint arXiv:2302. 12813* (2023).
200. Luo, L., Li, Y.-F., Haffari, G. & Pan, S. Reasoning on Graphs: Faithful and Interpretable Large Language Model Reasoning. (2023).
201. Ji, Z. *et al.* Towards Mitigating LLM Hallucination via Self Reflection. 1827–1843 (2023) doi:10.18653/V1/2023.FINDINGS-EMNLP.123.
202. Yao, S. *et al.* Tree of Thoughts: Deliberate Problem Solving with Large Language Models. *Adv Neural Inf Process Syst* **36**, (2023).
203. Ye, X. & Durrett, G. The Unreliability of Explanations in Few-shot Prompting for Textual Reasoning. *Adv Neural Inf Process Syst* **35**, (2022).
204. Ghandeharioun, A., Caciularu, A., Pearce, A., Dixon, L. & Geva, M. Patchscopes: A Unifying Framework for Inspecting Hidden Representations of Language Models. *Proc Mach Learn Res* **235**, 15466–15490 (2024).

205. Hyland, K. Academic publishing and the myth of linguistic injustice. *J Second Lang Writ* **31**, 58–69 (2016).
206. Clavero, M. '"Awkward wording. Rephrase"': linguistic injustice in ecological journals. (2010).
207. Strauss, P. Shakespeare and the English poets: The influence of native speaking English reviewers on the acceptance of journal articles. *Publications* **7**, 20 (2019).
208. Yanai, I. & Lercher, M. Night science. *Genome Biol* **20**, 1–3 (2019).
209. Silver, D. *et al.* Mastering the game of Go with deep neural networks and tree search. *Nature 2016 529:7587* **529**, 484–489 (2016).
210. Silver, D. *et al.* A general reinforcement learning algorithm that masters chess, shogi, and Go through self-play. *Science (1979)* **362**, 1140–1144 (2018).
211. Zhao, A. *et al.* Absolute Zero: Reinforced Self-play Reasoning with Zero Data. (2025).
212. Tegnér, J. N. *et al.* Computational disease modeling - Fact or fiction? *BMC Syst Biol* **3**, 1–3 (2009).